# Socially acceptable route planning and trajectory behavior analysis of personal mobility device for mobility management with improved sensing

Sumit Mishra[1], Praveen Kumar Rajendran[2], and Dongsoo Har[3]

[1] The Robotics Program, Department of Electrical Engineering, Korea Advanced Institute of Science and Technology, Daejeon, Republic of Korea

[2] Division of Future Vehicle, Department of Electrical Engineering, Korea Advanced Institute of Science and Technology, Daejeon, Republic of Korea

[3] The CCS Graduate School of Green Transportation, Korea Advanced Institute of Science and Technology, Daejeon, Republic of Korea

`sumitmishra209@gmail.com, praveenkumar@kaist.ac.kr,`
`dshar@kaist.ac.kr`

**Abstract.** In urban cities, with increasing acceptability of shared spaces used by pedestrians and personal mobility devices (PMDs), there is need for pragmatic socially acceptable path planning and navigation management policies. Hence, we propose a socially acceptable global route planner and assess the legibility of the resulting global route. Our approach proposed for choosing global route avoids streets penetrating shared spaces and main routes with high probability of dense usage. The experimental study shows that socially acceptable routes can be effectively found with an average of 10 % increment of route length with optimal hyperparameters. This helps PMDs to reach the goal while taking a socially acceptable and safe route with minimal interaction of different PMDs and pedestrians. When PMDs interact with pedestrians and other types of PMDs in shared spaces, micro-mobility simulations are of prime usage for acceptable and safe navigation policy. Social force models being state of the art for pedestrian simulation are calibrated for capturing random movements of pedestrian behavior. Social force model with calibration can imitate the required behavior of PMDs in a pedestrian mix navigation scheme. Based on calibrated models, simulations on shared space links and gate structures are performed to assist policies related to deciding waiting and stopping time. Also, based on simulated PMDs interaction with pedestrians, location data with finer resolution can be obtained if the resolution of GPS sensor is 0.2 m or less. This will help in formalizing better modelling and hence better micro-mobility policies.

# 1 Introduction

Mobility management is defined differently, depending on specific applications. It is for managing coordinated transportation services in transportation engineering, while it represents locating a mobile subscriber's point of attachment for delivering data packets in wireless communication network such as OFDM network [1]. With advanced communication technology handling micro-mobility in connected smart city is becoming a key component of future urban smart mobility. However, continued adoption of smart micro-mobility devices, such as electric skateboards, kick scooters, and self-balancing unicycles, uplifts queries about the safety and optimal administration of smart micro-mobility. A study that analyzed the rising injury events related to PMD devices in Singapore shows that about 15% of events required surgery and 6 deaths have occurred with motorized PMDs [2]. Many European countries, such as the Netherlands, Denmark, France, and Belgium, as well as a few Asian countries, such as Korea and Japan, support cycling-specific lanes. However, allotting specific lanes for every type of PMD is not possible and hence these countries also promote a shared space for mixed types of PMDs along with other micro-mobility modes like walking. The global personal transportation industry is anticipated to double in size between 2014 and 2023, as per Transparency Market Research [3]. Non-motorized transportation, on the other hand, allows people to get some exercise, which is beneficial to their health and environmental goals [4].

In urban cities shared spaces are gaining popularity. The ideology of shared space is to make better use of urban streets with pleasant behavior experience to users. The shared space can reduce isolation between different types of road users, making walkable and easily commutable cities [5]. Rather than pivoting to traffic segregation, city governing bodies drive towards mutual usage of the shared space [6]. The strategy of not having a distinct infrastructure for diverse modes of road users is unquestionably cost-effective. Users in shared space can enjoy riding small and lightweight vehicles which usually operate at a minimal pace.

The density and speed of micro-mobility are inversely related, whereas the density of micro-mobility and conflict incident rate is directly proportionate. A case study points out that mixed usage of PMDs in shared-space cause negative impact on pedestrian's choice of using that space. Hence, concern of pedestrians need to be considered for deploying socially acceptable routes and navigation in sustainable smart cities [7]. Different elements of the shared space and the interaction between them are crucial for regulating and devising administration proposal, designing an infrastructure, or policy formation [8]. This interpretation accounts for minimizing the density as much as possible and if not possible the measures to improve safety are considered by the behavior analysis of users of shared space [6]. Thus, path planning is needed to choose a path from the available paths for a global route, with the least penetration of PMDs in shared space. Qualitative discussion and policies on route choice and management of PMDs in heterogeneous traffic of non-motorized facility are presented in [4, 8, 9], however, in their works mathematical quantitative model for micro-level and global level route planning and management is missing. Micro-planning is necessary because PMDs are using shared space and navigation policies like flow density or stop and wait time are constructed in accord with pedestrian behavior. Simulation can be useful to decide the resolution of GPS sensor for finer location data. This will help to understand practical behavior in shared spaces [10] and will lead to better micro-policy formation. Micro-behavior of PMDs for better modelling is inspected in [5] and in [11] mathematical analysis using social force is presented. Still the inspection using full-fledge social force model is pending. From this point of view, the contributions of our work can be summarized as following:

- Design criteria and mathematical models for weight allotments of links which are most effective to achieve socially acceptable routes avoiding penetration in social space are described and map-based visualizations are provided.
- Global route analysis is made by test run performed with shared space and PMDs type-specific lanes in Delhi-NCR, India.
- Social force model and their implications for local mobility policy management of PMDs are introduced.
- Using simulation and the empirically noted calibration parameters, this study shows how socially acceptable pass-by time varies according to different types of PMDs.

Solutions devised in this work can be effectively used for making regulatory framework by city council to introduce PMDs in smart cities. Section 2 provides details of methods for finding socially acceptable global routes. A safe route finding methodology is also discussed. Section 3 presents optimal results and case study discussion for local navigation considering models of pedestrian behavior. Finally, a conclusion is drawn from the study followed by the references.

## 2  Socially Acceptable Global Path Analysis for PMD

### 2.1  Map visualization of shared space in an area

For global path analysis, we need to know availability of road types. These maps are usually maintained and updated by authorities, private organizations, and crowd-sourced data from a platform like open street maps. Shared types of road space can be divided into two, first type corresponds to 'meso-unavailability', e.g., unavailability of road types spanning in an area or zone. The other type is 'micro-unavailability' for a specific intersection or a road for which given road-types are not available. In a region of interest, we got list of shared spaces, zones of 'meso-unavailability' as well as list of point-specific intersection as 'micro-unavailability'. Information of zones for specific type of road availability is in the form of street ID hence meso-unavailability of a given type can be found by subtracting from the set of all the road street IDs.

Static visualization, that can be printed and shared on reports, as well as interactive visualization, which allows exploration of a data set, via zooming and toggling of overlays, are being made. Fig.1 shows visualization of road-type availability in Delhi, India. Three shades of color mapping are used for circular node markers with the highest color tone of pink for nodes within the inner 100 meters, middle color tone of pink for 100-200 meters and lowest color tone of pink for 200-300 meters of zone geocode point. The nodes within 100 meters proximity of specific intersection and road regions of 'micro-unavailability' of given road type are plotted with violet color, as shown in Fig. 1(a). For interactive visualization, Folium library is used in Fig. 1(b).

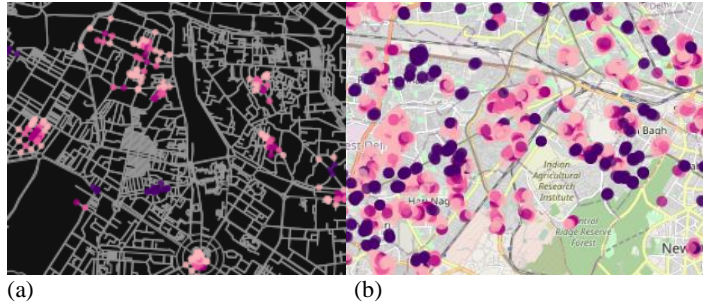

(a)          (b)

**Fig. 1.** Visualization of road-type availability in Delhi, India.

### 2.2 Traffic flow and Betweenness Centrality (BC) for finding out major links

The traffic flow in an edge majorly depends on the connectivity that an edge offers based on its presence in the different routes i.e. betweenness centrality. The Betweenness centrality ($c_B(e)$) of an edge is "proportional to the number of shortest paths between all pairs of nodes passing through it and can be measured by averaging over each pair of nodes and following the shortest path to the destination". High betweenness centrality corresponds to the edges of main roads and low is for small streets with low connectivity. Further, some studies show a high positive correlation between the traffic flow and the values of betweenness centrality of the road [12]. So, for searching a low connected street with less traffic flow, normalized edge betweenness centrality is a good parameter. There are inbuilt functions in Networkx that calculate the betweenness centrality of a given graph network [13].

$$c_B(e) = \sum_{s,t \in V} \frac{\sigma(s,t|e)}{\sigma(s,t)} \tag{1}$$

where V is the set of nodes, σ(s,t) is the number of shortest (s,t)-paths, and σ(s,t|e) is the number of those paths passing through edge e. For normalization, "1/(n(n−1))" factor is multiplied for directed graphs where n is the number of nodes in G.

### 2.3 Design criteria for selecting social acceptable route

The requirements of route planning for PMD are based on maximizing the usage of a given available road-type for a given PMD type and minimizing the usage of shared spaces for that type. Also, connectivity of roads needs to be considered for choosing the less connected roads with less flow as the more connected road has a high probability that road allotted for PMD will be used more by other PMDs types and may be pedestrian as well. We standardized the social acceptability of a given route for PMD by checking the extent of penetration in the shared area, the more it is the less socially acceptable is the route. After having the shared space gradient visualized we would require the source and destination coordinates to calculate the route. The efficiency of a route is also taken into account by choosing the shortest path while considering the shared space to the destination to minimize PMD travelling time for saving energy consumption. The factors we consider are the penetration proximity to shared space zones. The edges present within the 100 meters radius of the zones center are valued as the most unfavorable edges to pass by, in the context of the study. The social-hazard score (HAZ) decreases with every 100 meters as the distance increases up to 300 meters (maximum extent) and is taken as 0.20, 0.16, and 0.12. Another social-hazard factor considered is proximity to 'micro-unavailability' type shared space. For that, we allotted the social-hazard score (PA) of 0.08 for the edges with the proximity of 100 meters as they are small shared spaces so are less used in comparison to meso-unavailability shared space type. Sometime for PMDs wireless charging is primal using installed wireless power station. For that, destination need to be set according to constraint of available battery charge in PMDs, for nearby charging station. As such whenever required, new route can be searched by changing optimal destination. Sequential path planning for optimal spot in next step for maximized objective has been a hot topic for various applications [14, 15].

Streets are usually less occupied by local pedestrians and different PDEs compared to main road pathways. So normalized edge betweenness centrality (BC) is considered as a weighted score for navigational traffic risk of global path. BC is normalized using a min-max scalar to lie in the range of 0 to 0.06. Another important requirement of the route is that it should not be much longer than that of the shortest route. For that to consider different hazards and betweenness centrality weight should be an empirically set factor of shortest path length. Other factors like spatio-temporal pedestrian or PDEs density can also be considered related to pedestrian and PDEs traffic if a database is available. Usually, such databases are created passively by historically memorizing the pedestrian and PDEs usage density on a road. Fig.2 shows block diagram consisting of each component as discussed for selection criteria.

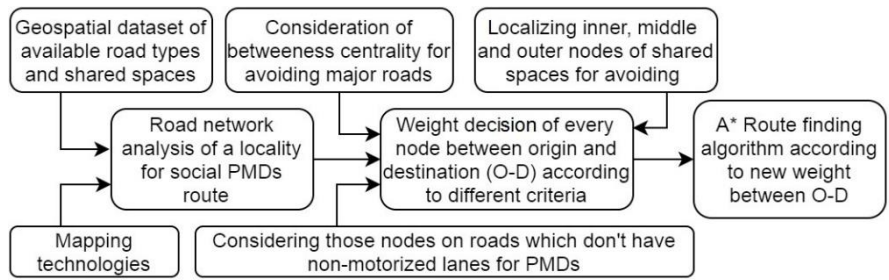

**Fig. 2.** Block diagram for social route selection criteria

### 2.4 Cost function of links for global path search

There are many algorithms to find shortest route, such as Dijkstra, Bellman-Ford, A*, etc. These routing algorithms need edge weight as input for every edge along the route to minimize the summation of weights for finding out the best optimal route. For our purpose, we use A* algorithm [16]. The edge weight should be set in a way that after summation of all the edge weights along the route do not undermine the social-acceptability effect of any given edge in the route. For this, social-acceptability effect of any given edge should be comparable to the edge length of the total route. Further different types of social-hazard scores should be comparable so that one does not undermine the effect of others, all together.

We allot the Edge weight of each edge as per equation (2). For finding out the edge weight a comparable constant to route length is taken which is weighted by the total social-hazard score. The comparable constant, which we used, is the shortest path length between the origin and destination point. Further, to calculate normalized centrality of edges, we take a sub-network bounded by a square bounding box of side 5km. This ensures that centrality calculation does not take account of those routes that are remote to any probable path between origin and destination. The hazard scores are set in a way to give the most importance to inner zone area and specific intersection points. Once the edge weight of sub-network has been updated, the shortest path finding algorithm will search for an optimal path using edge weight given by

Edge weight= Edge length + ((HAZ + PA + BC)* Shortest path length between origin and destination)  (2)

where HAZ is zone social-hazard score of edges, PA is the specific intersection point weight of joining edges and BC is the normalized betweenness centrality of the sub-network. Further, if the passive pedestrian and PMDs usage density (IC) database are available, then for decreasing interaction of PMD, this can also be considered as in equation (3). However, IC should be used only after normalization to a comparable range of values with other weights. Accordingly edge weight can be given by

Edge weight= Edge length + ((HAZ + PA + BC+IC)* Shortest path length between origin and destination)  (3)

### 2.5 Global path analysis

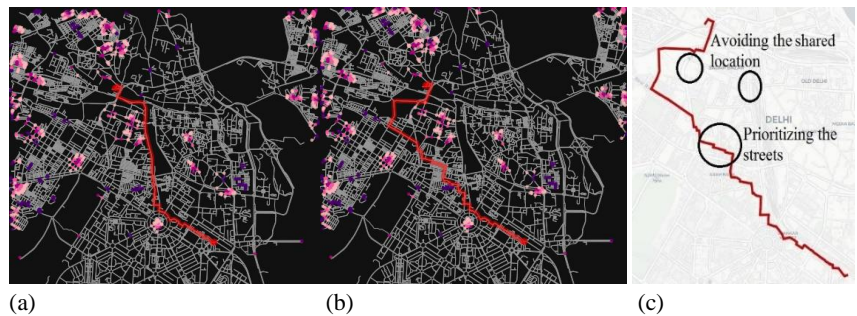

(a)  (b)  (c)

**Fig. 3.** (a) Shortest route for a given O(rigin)-D(estination) pair (b) New route for the same O-D pair (c) Highlight of advantages of the new route for a given O-D pair.

As discussed, the required route should not be much longer than the shortest route. It should also avoid shared spaces and specific intersection points and avoid main roads.

Empirically HAZ, PA and BC should be allotted values such that if some social unacceptability is available on all possible routes then it should select the route which is relatively more acceptable and also not too long as compared to the shortest path. While selecting a route, it may go from less socially acceptable points if the other possible route is too long in comparison with the prior. To analyze and test the proposed edge weight, we find the route between 1000 origin-destination (O-D) pairs in the Delhi region. These O-D pairs are chosen such that the Euclidean distance between them is between 4.5 km and 6.5 km. The A* optimal route finding algorithm mainly employed by the majority of navigational systems is used. The Average length of 1000 short routes, as well as the average length of 1000 new-routes according to the proposed algorithm, is noted in Table 1, along with percentage change. Hence in general the proposed algorithm outputs nearly comparable route lengths to that of the shortest route with an extension near around 10%. Furthermore, for a given O-D pair, Fig. 3(a) and Fig. 3(b) show shortest path as well as new path. Fig. 3(c) highlights that the proposed algorithm prioritizes the selection of streets over main roads. After the empirical study with 1000 different routes, we found that hyper-parameters of column 2 in Table 1 are most viable and optimal, considering both hazard and length.

Table 1. Cost function hyperparameters and the related average of 1000 different path lengths

|   | HAZ (for 0-100m) | HAZ (for 100-200 m) | HAZ (for 200-300 m) | PA | BC (max range) | Avg. of shortest paths | Avg. of new path | % increment |
|---|---|---|---|---|---|---|---|---|
| 1 | 0.5 | 0.4 | 0.3 | 0.2 | 0.15 | 7161.2 | 8445.3 | 17.9 |
| 2 | 0.2 | 0.16 | 0.12 | 0.08 | 0.06 | 7161.2 | 7814.9 | 9.12 |
| 3 | 0.3 | 0.24 | 0.18 | 0.12 | 0.09 | 7161.2 | 8029.1 | 12.1 |

## 3 Socially Acceptable Local Path Management for PMD

### 3.1 Modelling of socially acceptable local navigation

Socially acceptable local navigation in shared space can be considered an active approach to promote micro-mobility of pedestrians and find out the best optimal policy for management. Management like bottleneck flow decisions, by managing stops and weight time leverages simulations for navigation modelling. Further various PMDs can use the shared space if the navigation model is designed to consider the required speed and space for different PMDs. This navigation should act in a way that PMD should grant rights-of-way to other types of PMDs as well as pedestrians in, but also prevent from being immobilized for an extended period like that of human's navigation scheme. A shy distance normally observed in human's navigation can also be maintained. Hence to incorporate this behavior in simulations, real-world pedestrian model is needed for developing a socially acceptable navigation scheme of PMDs. In a human crowd walking characteristic decisions are based on the individual as well as group characteristics which is imposed on PMDs too.

Individual characteristics are mathematically expressed as follows
- The attractive forces between each pedestrian and their destination point is

$$\vec{f}_i^0 = \frac{d\vec{v}_i}{dt} = \frac{v_i^0 \vec{e}_i^0 - \vec{v}_i(t)}{\tau} \tag{4}$$

where $\vec{v}_i(t)$ is the velocity of i-th pedestrian, with the maximum desired speed $v_i^0$ (~1.3m/s) and a desired direction of motion $\vec{e}_i^0$ within a certain relaxation time $\tau$ (=0.5 s) [17].

- Repulsive forces from obstacles is an exponentially decaying function of the distance $d_w$ perpendicular to the boundary is

$$\overrightarrow{f_i^{\text{wall}}}(d_w) = ae^{-d_w/b} \tag{5}$$

where, $a$=10 ms$^{-2}$ and b=0.1 m, and repulsion approximately extends over 30 cm [18].

- The repulsive forces between pedestrians is a negative differential of exponentially decaying potential defined by elliptical equipotential boundaries in moving direction as shown in Fig. 4 and is given by

$$\vec{f}_{\alpha\beta}(\vec{r}_{\alpha\beta}) = -\nabla_{\vec{r}_{\alpha\beta}} V_{\alpha\beta}[b(\vec{r}_{\alpha\beta})], \tag{6a}$$

$$V_{\alpha\beta}(b) = V_{\alpha\beta}^0 e^{-\frac{b}{\sigma}}, \tag{6b}$$

$$2b = \sqrt{\left(\|\vec{r}_{\alpha\beta}\| + \|\vec{r}_{\alpha\beta} - v_\beta \Delta t \vec{e}_\beta\|\right)^2 - \left(v_\beta \Delta t\right)^2} \tag{6c}$$

where, $\vec{r}_{\alpha\beta} := \vec{r}_\alpha - \vec{r}_\beta$, step width ($s_\beta = v_\beta \Delta t$) of pedestrian, $V_{\alpha\beta}^0$= 2.1 m$^2$ s$^{-2}$, $\sigma$= 0.3m is incorporated from [19].

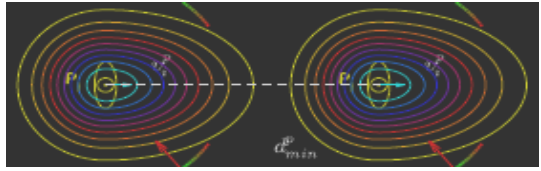

**Fig. 4.** Force field in the direction of movement for pedestrian

Group Characteristics are mathematically expressed as follows
- The coherence force that holds group members together

$$\vec{f}_i^{att} = q_A \beta_2 \vec{U}_i \tag{7}$$

where, $\beta_2$ is attractive strength in $\vec{U}_i$ unit vector direction pointing from pedestrian i to group center of mass. $q_A$ is 1 or 0 based on threshold distance between pedestrian i and the group's COM

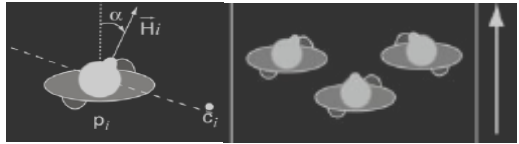

**Fig. 5.** Angle between gaze and movent direction of group formation

- A force calculated from the gaze directions of pedestrians to maintain group formations.

$$\vec{f}_i^{vis} = -\beta_1 \alpha_i \vec{V}_i \tag{8}$$

where, the negative sign represents greater $\alpha_i$ leads to less comfortable in the turning for walking as shown in Fig. 5. $\beta_1$ describes social interactions strength between group members in $\vec{V}_i$ the velocity vector of pedestrian i.

- Repulsive force that keeps members from getting too close to each other is

$$\vec{f}_i^{rep} = \sum_k q_R \beta_3 \vec{W}_{ik} \tag{9}$$

where, $\beta_3$ is repulsion strength in $\vec{W}_{ik}$ unit vector direction pointing from pedestrian i to the group member k, and $q_R$ is 1 or 0 based on threshold distance between pedestrians 'i' and 'k'. Values of constant parameters for group characteristics are adopted from [20]. Combining individual and group forces, total force on a pedestrian is:

$$\frac{d\vec{v}_i}{dt} = \vec{f}_i^0 + \vec{f}_i^{wall} + \sum_j \vec{f}_{ij} + \vec{f}_i^{group} \tag{10}$$

### 3.2 Local navigation analysis

**Table 2.** Parameters for force in SFM models for PMDs that are similar to pedestrian and for the PMDs that are faster and more space taking than that of pedestrian.

|  | Goal Attractive force factor | Pedestrian repulsive force factor | Space repulsive force factor | Social force factor | Obstacle force factor |
|---|---|---|---|---|---|
| PMDs with low speed and less space (Type-1) | 1 | 1.5 | 1 | 5.1 | 10 |
| PMDs with high speed and more space (Type-2) | 0.7 | 3.0 | 2.1 | 6.6 | 18 |

In Table 2, each type of forces can be explained as follows:
- Goal Attractive Force: Force responsible for accelerating to desired velocity
- PMD Repulsive Force: Force responsible for PMD to PMD repulsive force
- Space Repulsive Force: Force responsible for Obstacles to PMD repulsive force
- Social Force: Force between a given PMD and all other PMD belong to same scene
- Obstacle Force: Force between a given PMD and the nearest obstacle in the scene.

**Table 3.** Simulations end time in seconds based on two different sets of parameters of micro-mobility in Table 2

| Simulation | Type-1 | Type-2 |
|---|---|---|
| 1. Gate structure-low crowd | 67 | 88 |
| 2. Street-low crowd | 42 | 51 |
| 3. Street-heavy Crowd | 50 | 55 |

Using the resultant force field, different simulations with different PMDs counts, type, starting and end-goal points, moving direction, and obstacles are conducted. In the first simulation, we generate two PMDs going in the opposite directions with two obstacles. In the second simulation, there are three PMDs with two obstacles. Trajectories are smooth for less number of pedestrians and the divergence in trajectory is there when they encounter obstacles. However, when the two PMDs interact, the trajectory is a little complex and has high similarity with the real word human trajectory as shown in Fig. 6 (a). Further, to simulate a crowded situation, we create five PMDs, all independently going to their targets. Then we make one group of two PMDs in that simulation for five PMDs. The trajectory as traced in Fig. 6 (b) is complex and chaotic similar to human movements in crowded conditions. Hence the resultant force field is capable of generating socially acceptable navigation trajectory in sparse as well as dense PMD conditions. Using the same socially acceptable PMD navigation models, flow management scheme can be deployed in shared urban space.

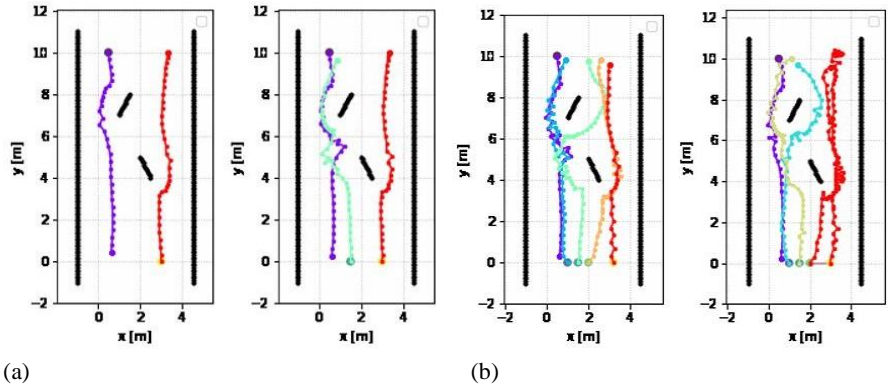

(a)  (b)

**Fig. 6.** (a) Simulation for low crowd condition and (b) Simulation for high crowd condition (Trajectory of PMDs are shown with different color unless two PMDs are in a group. Also starting point is showen by big circle than the end point of the traced trajectory.)

We performed three different simulations to predict crowd behavior for three different situations – driving on a street with mixed type of PMDs in the low and heavy crowd, and ingress and egress through the gate structure. We calculated the time to end the simulation using the different parameters mentioned in Table 2 and then tabulated the time in seconds for two types of PMDs in comparison with pedestrians, in Table 3. The parameter for SFM based PMD models as enlisted in Table 2 are empirical. Exact analysis of these parameters can be done for different types of PMDs in experiments with human subjects as done in [11]. However, all the factors of the advanced SFM model have not been considered in that study. Also, the parameters so calculated can't be taken as standard as every shared space does have flexibility toward the social acceptance of PMDs. The number of seconds represents the time taken to complete the similar goals as enlisted for 3 simulations and is increased for second type of PMD for all the mentioned situations (see Table 3), which is acceptable social behavior that means high speed PMDs with more area should be slower at shared space. The complex trajectory changes in simulation is observed to have minimum distance of 0.2 meters in consecutive samples. Hence, if the resolution of GPS sensing device is equal to or less than 0.2 meters then finer real-world trajectory-data can be captured for better micro-policy. Hence empirical parameters of SFM models can be recalibrated for location specific tracing of PMDs in urban space.

## 4      Conclusion

Proposed schemes implement vital solution for PMD mobility management for global as well as local prospects in urban cities with shared spaces. The paper lays down some facts on the usage of type-specific as well as mix type roads wherever available for socially acceptable PMD mobility. This pilot study has proven that the system works adequately for both local and global mobility policy management. The present work establishes algorithmic approaches to deduce global paths for the socially acceptable mobility of PMD. This paper presents the technicality of our objective function to assist in socially acceptable routes. Further, for local micro-mobility policies, simulation models are developed similar to human walking patterns and studied using simulation runs of different scenarios. The simulation with calibrated model establishes that the PMDs with high speed and more area are taking more time in comparison to low speed and smaller size PMDs, which can be taken as socially acceptable. Also based on complex simulated trajectory, if the resolution of GPS sensor on PMD is less than 0.2 m then more accurate hyperparameters for real-time trajectory behavior in SFM model can be analyzed from the location data. For global path, this work includes test cases run at several locations in Delhi, India with an average 10 % increment of route length with optimal hyperparameters. In the case of local management, using simulation, the study shows how pass-by time can vary for different PMDs. Hence policymakers can use it to establish PMDs flow density and stop and pass-by times for encouraging mixed micro-mobility in urban cities at shared spaces.

## References

[1] H. Kim, E. Hong, C. Ahn, and D. Har, "A pilot symbol pattern enabling data recovery without side information in PTS-based OFDM systems," *IEEE Transactions on Broadcasting,* vol. 57, no. 2, pp. 307-312, 2011.

[2] A. L. Tan, N. Nadkarni, and T. H. Wong, "The price of personal mobility: burden of injury and mortality from personal mobility devices in Singapore-a nationwide cohort study," *BMC public health,* vol. 19, no. 1, pp. 1-7, 2019.

[3] G. W. Shin, K.-J. Lee, D. Park, J. H. Lee, and M. H. Yun, "Personal Mobility Device and User Experience: A State-of-the-art Literature Review," in *Proceedings of the Human Factors and Ergonomics Society Annual Meeting*, 2018, vol. 62, no. 1: SAGE Publications Sage CA: Los Angeles, CA, pp. 1336-1337.

[4] T. Litman and R. Blair, *Managing personal mobility devices (PMDs) on nonmotorized facilities*. Victoria Transport Policy Institute, 2017.

[5] M. Che, Y. D. Wong, K. M. Lum, and X. Wang, "Interaction behaviour of active mobility users in shared space," *Transportation Research Part A: Policy and Practice,* vol. 153, pp. 52-65, 2021.

[6] D. Beitel, J. Stipancic, K. Manaugh, and L. Miranda-Moreno, "Assessing safety of shared space using cyclist-pedestrian interactions and automated video conflict analysis," *Transportation research part D: transport and environment,* vol. 65, pp. 710-724, 2018.

[7] B. Ruiz-Apilánez, K. Karimi, I. García-Camacha, and R. Martín, "Shared space streets: design, user perception and performance," *Urban Design International,* vol. 22, no. 3, pp. 267-284, 2017.

[8] S. Akter, M. M. H. Mamun, J. L. Mwakalonge, G. Comert, and S. Siuhi, "A policy review of electric personal assistive mobility devices," *Transportation research interdisciplinary perspectives,* vol. 11, p. 100426, 2021.

[9] N. T. Commission, "Barriers to the safe use of personal mobility devices. 2019.[Cited 26 Nov 2019.]," ed.

[10] C. Butron-Revilla, E. Suarez-Lopez, and L. Laura-Ochoa, "Discovering Urban Mobility Patterns and Demand for Uses of Urban Spaces from Mobile Phone Data," in *2021 2nd Sustainable Cities Latin America Conference (SCLA)*, 2021: IEEE, pp. 1-6.

[11] Y. Hasegawa, C. Dias, M. Iryo-Asano, and H. Nishiuchi, "Modeling pedestrians' subjective danger perception toward personal mobility vehicles," *Transportation research part F: traffic psychology and behaviour,* vol. 56, pp. 256-267, 2018.

[12] Y. Liu, X. Wei, L. Jiao, and H. Wang, "Relationships between street centrality and land use intensity in Wuhan, China," *Journal of Urban Planning and Development,* vol. 142, no. 1, p. 05015001, 2016.

[13] U. Brandes, "On variants of shortest-path betweenness centrality and their generic computation," *Social Networks,* vol. 30, no. 2, pp. 136-145, 2008.

[14] C. Moraes, S. Myung, S. Lee, and D. Har, "Distributed sensor nodes charged by mobile charger with directional antenna and by energy trading for balancing," *Sensors,* vol. 17, no. 1, p. 122, 2017.

[15] C. Moraes and D. Har, "Charging distributed sensor nodes exploiting clustering and energy trading," *IEEE Sensors Journal,* vol. 17, no. 2, pp. 546-555, 2016.

[16] J. Yao, C. Lin, X. Xie, A. J. Wang, and C.-C. Hung, "Path planning for virtual human motion using improved A* star algorithm," in *2010 Seventh international conference on information technology: new generations*, 2010: IEEE, pp. 1154-1158.

[17] D. Helbing and P. Molnar, "Social force model for pedestrian dynamics," *Physical review E,* vol. 51, no. 5, p. 4282, 1995.

[18] A. Johansson, D. Helbing, and P. K. Shukla, "Specification of the social force pedestrian model by evolutionary adjustment to video tracking data," *Advances in complex systems,* vol. 10, no. supp02, pp. 271-288, 2007.

[19] M. Moussaïd, D. Helbing, S. Garnier, A. Johansson, M. Combe, and G. Theraulaz, "Experimental study of the behavioural mechanisms underlying self-organization in human crowds," *Proceedings of the Royal Society B: Biological Sciences,* vol. 276, no. 1668, pp. 2755-2762, 2009.

[20] M. Moussaïd, N. Perozo, S. Garnier, D. Helbing, and G. Theraulaz, "The walking behaviour of pedestrian social groups and its impact on crowd dynamics," *PloS one,* vol. 5, no. 4, p. e10047, 2010.